\pdfoutput=1

\documentclass[11pt,dvipsnames]{article}

\usepackage[]{acl}

\usepackage{times}
\usepackage{latexsym}
\usepackage{amsmath}
\usepackage{amssymb}
\usepackage{booktabs}
\usepackage{cleveref}
\usepackage{graphicx}
\usepackage{xcolor}
\usepackage[normalem]{ulem}
\usepackage{mathtools}
\usepackage{microtype}

\usepackage{soul}

\usepackage[T1]{fontenc}

\usepackage[utf8]{inputenc}

\usepackage{microtype}

\title{Instilling Type Knowledge in Language Models via Multi-Task QA}

\author{Shuyang Li \\
  UC San Diego \\
  \texttt{shl008@ucsd.edu} \\\And
  Mukund Sridhar \\
  Amazon Alexa AI \\
  \texttt{harakere@amazon.com} \\\And
  Chandana Satya Prakash \\
  Amazon Alexa AI \\
  \texttt{chanprak@amazon.com} \\\AND
  Jin Cao \\
  Amazon Alexa AI \\
  \texttt{jincao@amazon.com} \\\And
  Wael Hamza \\
  Amazon Alexa AI \\
  \texttt{waelhamz@amazon.com} \\\And
  Julian McAuley \\
  UC San Diego \\
  \texttt{jmcauley@ucsd.edu}
 }

\begin{document}
\maketitle
\begin{abstract}
Understanding human language often necessitates understanding entities and their place in a taxonomy of knowledge---their \emph{types}.
Previous methods to learn entity types rely on training classifiers on datasets with coarse, noisy, and incomplete labels. 
We introduce a method to instill fine-grained type knowledge in language models with text-to-text pre-training on type-centric questions leveraging knowledge base documents and knowledge graphs.
We create the \textbf{WikiWiki} dataset: entities and passages from 10M Wikipedia articles linked to the Wikidata knowledge graph with 41K types.
Models trained on WikiWiki achieve state-of-the-art performance in zero-shot dialog state tracking benchmarks, accurately infer entity types in Wikipedia articles, and can discover new types deemed useful by human judges.
\end{abstract}

\section{Introduction}

Entities can be categorized by their \emph{types}, which indicate where they belong in a taxonomy of knowledge.
For example, Venus is a planet and 
thus also an astronomical body.
Much like how knowledge acquisition in cognitive development progresses from recognizing concrete objects to gradually understanding their relations to one another \cite{cogdev},
we aim to extend language models' existing rough understanding of entities \cite{LMent} to the types that govern how entities are related.
Instilling type knowledge in multi-purpose models can improve performance in 
tasks like entity linking \cite{interpET},  question-answering \cite{typesQA}, and semantic parsing \cite{semparse}.

While language models can memorize some facts \cite{LMKG}, they frequently hallucinate false information
\cite{kglm,LMhallucinate}.
Current attempts to learn to infer types for entities
are hampered by 1) the difficulty of collecting diverse, large-scale typing datasets; and 2) how existing corpora assume independence between types \cite{ultrafine}, while in reality types sit at levels of granularity that are useful in different settings:
a pizza store may care whether a user likes Cheese Pizza;
a restaurant recommender might care if the user wants Pizza;
finally, a general dialog agent might only care if a user wants Food.

\begin{figure}[t!]
    \centering
    \includegraphics[width=1.0\linewidth]{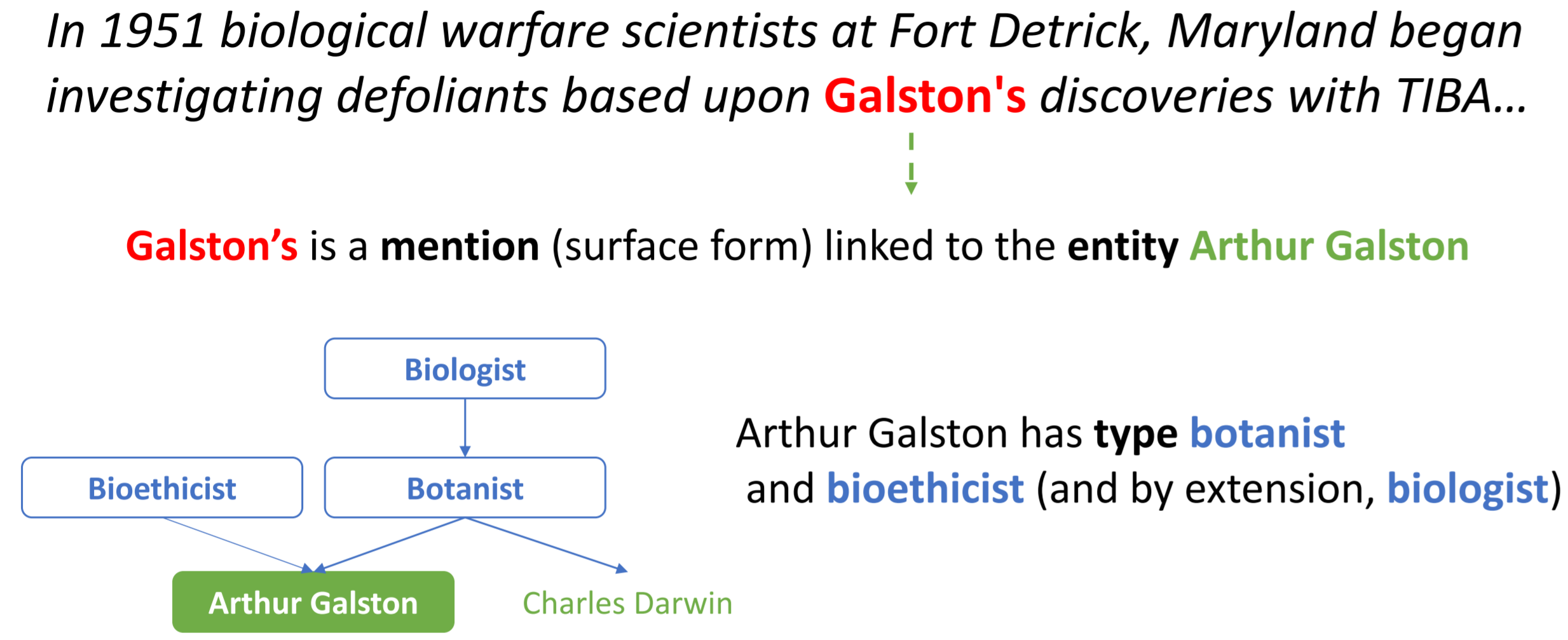}
    \caption{Via the \textbf{WikiWiki} dataset, we train a model to answer questions about entities mentioned in Wikipedia articles (top) and WIkidata \emph{types} that such entities are an 
    \emph{instance of} (P31) or \emph{subclass of} (P279).}
    \label{fig:definitions}
\end{figure}

We address both issues by proposing a simple and effective approach for pre-training generative language models to answer questions about entities, types, and surface forms (mentions) in a large public knowledge graph (KG) consisting of Wikipedia articles and Wikidata nodes.
We leverage
high quality type labels in a large corpus of knowledge-rich text \emph{and}
an ordered, hierarchical type ontology.

To summarize our main contributions:
1) We create the new \textbf{WikiWiki} dataset comprising 10M Wikipedia articles linked to nodes from Wikidata;
2) We propose a pre-training scheme for generative language models using type-centric question-answering based on WikiWiki;
3) We achieve state-of-the-art (SOTA) performance in zero-shot domain adaptation for dialog state tracking using our type-instilled models, with average per-domain gains of 14.9\% (49.4\% relative) joint accuracy;
and 4) We show that our models can precisely infer types for seen and unseen entities in new articles from WikiWiki, and propose novel types that humans judge to be accurate and appropriate.

\begin{table}[t!]
\centering
\small
\begin{tabular}{@{}lrrr@{}}
\toprule

                  & Training & Test & Test (New Ent) \\ \midrule
Documents         & 10 M   & 5.0 K   & 5.0 K \\
Unique Entities   & 2.2 M  & 14.1 K    & 6.0 K  \\
Unique Types      & 40.6 K   & 4.0 K   & 1.2 K \\
Num. of Mentions     & 38.7 M   & 19.3 K    & 6.4 K  \\
Type References   & 43.8 M   & 21.5 K    & 6.5 K  \\ \bottomrule
\end{tabular}
\caption{
Unique documents/entities/types and number of mentions in each split of \textbf{WikiWiki}. \emph{Test (New Ent)} comprises entities not seen in the training split.
}
\label{tab:dataset-statistics}
\end{table}

\section{Related Work}

\paragraph{Knowledge Grounding in Language Models}
Large pre-trained language models have been shown to memorize some facts \cite{LMKG}.
One recent line of work aims to explicitly condition generation on knowledge bases by combining a retrieval module and a language model \cite{interview,realm,rag,personachat}. 
\citet{knowbert} propose instead to align token representations from pre-trained language models with entity embeddings to reason over a limited set of entities.
\citet{luke} explicitly denote entity tokens with a learned input embedding.
Specific entity embeddings have also been learned jointly by using knowledge graphs as auxiliary inputs during language model pre-training \cite{colake,EAE,dkplm}.
Another line of work aims to model specific factual statements from knowledge bases \cite{kepler} for reading comprehension \cite{kelm} and trivia QA \cite{tekgen}.
We propose
text-to-text pre-training on knowledge recovery tasks to instill \emph{type-awareness}.
Our models
learn type knowledge that transfers to the type-adjacent downstream task of dialog state tracking and can infer unseen types.

\paragraph{Entity Representation Learning}
Many SOTA systems for knowledge retrieval and QA rely on learned dense embeddings of individual entities or types to perform multi-class classification \cite{ganeaNED,DPRQA,ZSEL}.
Several recent frameworks aim to learn entity knowledge during language model pre-training via entity masking \cite{ernie} or contrastive learning \cite{erica}.
Systems for entity typing \cite{daiET} and disambiguation \cite{yamadaED} also learn dense vector encodings 
that are later matched via dot-product scoring.
\citet{genre} aim to address some downsides of the above approaches---the linearly increasing space required to store learned representations and 
difficulties in negative sampling---by casting the task as generative language modeling: predict the name of an entity to be linked.
We generalize this approach from entity names (which appear verbatim) to include types, which require a more nuanced understanding of a context.

\section{Type-Centric Multitask Modeling}
\paragraph{WikiWiki Corpus}

To train an entity- and type-aware language model, we build the \textbf{WikiWiki} dataset by combining Wikipedia articles with the Wikidata KG \cite{Wikidata}.
Wikipedia articles have been used to enrich
corpora for 
dialog \cite{wow},
coreference resolution \cite{wikilinks},
and QA \cite{rikinet}.
KGs
have been used for entity typing and relation extraction \cite{falcon}.
\citet{docred} use Wikipedia pages as context for relation triples mined from Wikidata.




\begin{table}[t!]
\centering
\small
\begin{tabular}{@{}l@{}}
\toprule
\begin{tabular}[c]{@{}l@{}}\textbf{Context:} These included carbon dioxide by burning\\ diamond, and \textcolor{OliveGreen}{mercuric oxide} by heating \textcolor{OliveGreen}{mercury}. This\\ type of experiment contributed to the discovery of\\ ``\textcolor{OliveGreen}{dephlogisticated air}'' by \textcolor{OliveGreen}{Priestley}, which became better\\ known as oxygen, following \textcolor{OliveGreen}{Lavoisier}'s investigations.\end{tabular} \\ \midrule
\begin{tabular}[c]{@{}l@{}}\textbf{Entity/Type Discovery (20\%):} List all concepts and\\ types mentioned here.\end{tabular} \\
\begin{tabular}[c]{@{}l@{}}\textbf{Answer:} \textcolor{OliveGreen}{Priestley} (\textcolor{RoyalBlue}{chemist}), \textcolor{OliveGreen}{Lavoisier} (\textcolor{RoyalBlue}{chemist}), \textcolor{OliveGreen}{mercuric}\\ \textcolor{OliveGreen}{oxide} (\textcolor{RoyalBlue}{chemical compound}), \textcolor{OliveGreen}{mercury} (\textcolor{RoyalBlue}{chemical element}),\\ and \textcolor{OliveGreen}{dephlogisticated air} (\textcolor{RoyalBlue}{superseded scientific theory})\end{tabular} \\ \midrule
\begin{tabular}[c]{@{}l@{}}\textbf{Entity Typing (30\%):} What is \textcolor{OliveGreen}{dephlogisticated air} an\\ example of?\end{tabular}         \\
\textbf{Answer:} \textcolor{RoyalBlue}{superseded scientific theory}                                                                                \\ \midrule
\textbf{Entity Recognition (20\%):} What does \textcolor{OliveGreen}{Priestley} refer to?                                                           \\
\textbf{Answer:} \textcolor{Red}{Joseph Priestley} (\textcolor{RoyalBlue}{chemist})                                                                                  \\ \midrule
\textbf{Slot Filling (30\%):} Which \textcolor{RoyalBlue}{chemists} are mentioned here?                                                             \\
\textbf{Answer:} \textcolor{Red}{Joseph Priestley} and \textcolor{Red}{Antoine Lavoisier}                                                                      \\ \bottomrule
\end{tabular}
\caption{In pre-training, the model reads a Wikipedia article and answers questions from four tasks involving entities and types. It must generate answers containing terms not found verbatim in the text. Surface forms (mentions) in \textcolor{OliveGreen}{\textbf{green}}, entities in \textcolor{Red}{\textbf{red}}, and types in \textcolor{RoyalBlue}{\textbf{blue}}.}
\label{tab:pretrain-ex}
\end{table}

We link articles, entities, and types as in \Cref{fig:definitions}:
like \citet{blink}, we take Wikipedia hyperlinks as links between \emph{entities}
(target
page) and their \emph{mentions} (link text);
we link pages to Wikidata nodes via ID;
and for each node we extract types $T$ from Wikidata where $t\in T$
is an \emph{instance}/\emph{subclass} of the node (discarding entities with no types).\footnote{All humans on Wikidata are an instance of `human'; we thus use the `occupation' relation to determine their types.}
To address sparsity of hyperlinks, we follow \citet{docred} and
use spaCy to identify additional entities.
We sample 10M articles for training, with two disjoint 5K-article splits for evaluation, containing seen and unseen (New Ent) entities respectively (\Cref{tab:dataset-statistics}).
The ontology of Wikidata types forms a directed acyclic graph with 41K type nodes applying to 2.2M entities.
Previous entity typing datasets rely on annotations from small groups of crowd-workers and include a small type ontology in the hundreds \cite{figer} and/or sacrifice label accuracy \cite{ultrafine}.
We instead rely on the cumulative, cross-checked annotations from tens of thousands of active Wikidata users.

Entities in Wikidata on average are assigned 1.28 types; for entities with multiple types, not all types are necessarily relevant to a context.
For example, take the following passage: \emph{``Obama was elected to the Illinois Senate in 1996, succeeding Democratic State Senator Alice Palmer from Illinois's 13th District, which, at that time, spanned Chicago South Side neighborhoods from Hyde Park–Kenwood south to South Shore and west to Chicago Lawn.''}

While Wikidata entities may have 5+ types, many are not directly relevant to a context.
For example, while Barack Obama has types including \emph{Politician}, \emph{Jurist}, \emph{Political Writer}, \emph{Community Organizer}, and \emph{Podcaster}, the latter is not relevant to the context.
To teach our models to infer types relevant to the context at hand, in pre-training data we take only types that are shared between Barack Obama and other entities in the document (e.g.~Alice Palmer---Politician).
We have made the WikiWiki dataset publicly available on Github.\footnote{\url{https://github.com/amazon-research/wikiwiki-dataset/}}


\paragraph{Pre-training Tasks}
To instill type-centric knowledge from WikiWiki, we train our models to answer four types of knowledge-based questions conditioned on a passage from Wikipedia (examples in \Cref{tab:pretrain-ex}).
In \emph{entity/type discovery}, the model is tasked to recover all surface forms (mentions) that reference an entity, along with their types---this is analogous to simultaneous entity recognition and typing.
\emph{Entity typing} consists of assigning types to an entity of interest.
For \emph{entity recognition}, we follow \citet{genre} by training our model to respond with an entity's full name and type when queried with a surface form.
In \emph{slot filling} we ask our model to return all entities mentioned in the passage belonging to a certain type.
For multi-type entities, we use a subset of relevant types given other entities in the context (\Cref{sec:appendix-data}). 
We treat QA as a universal format for diverse NLU tasks \cite{decaNLP}, and adopt the framework of \citet{t5} to treat each of our tasks as text-to-text generative modeling.
We create 50M questions for pre-training.

\paragraph{Model Architecture}
We use an encoder-decoder \cite{seq2seq} model initialized from BART---a Transformer \cite{transformer} language model pre-trained via de-noising autoencoding \cite{bart}.
Our model generates an answer $a$ as a text sequence given a document $D$ of length $t_d$ and question $q$.
The document is encoded via the encoder---consisting of $l$ Transformer layers of hidden dimensionality $h$, each applying 16-headed self-attention---to produce
$z \coloneqq \text{Enc}(D) \in \mathbb{R}^{t_d \times h}$.

\begin{table}[t!]
\small
\begin{tabular}{@{}ll@{}}
\toprule
User:         & \begin{tabular}[c]{@{}l@{}}I'm looking for a place to stay during my \\ upcoming trip to Cambridge.\end{tabular}                                                  \\ \midrule
System:       & \begin{tabular}[c]{@{}l@{}}I can definitely help you with that! What \\ area are you staying in, and what is the \\ price range you are looking for?\end{tabular} \\ \midrule
User:         & \begin{tabular}[c]{@{}l@{}}It should be located in the west and it\\ should be cheap.\end{tabular}                                                                \\ \midrule
\textbf{Belief State:} & \textcolor{red}{{[}hotel price range{]}}: \textcolor{blue}{cheap}; \textcolor{red}{{[}hotel area{]}}: \textcolor{blue}{west}                                            \\ \bottomrule
\end{tabular}
\caption{In Dialog State Tracking (DST), a model infers the belief state of a user given the dialog history thus far, comprising slots (\textcolor{red}{red}) and their values (\textcolor{blue}{blue}). In Zero-shot DST, the model must infer the correct values for slots that it has not seen during training, requiring the agent to rely on general type knowledge.}
\label{tab:dst-example}
\end{table}

We train the model to perform QA via conditional language modeling.
Instead of concatenating the question with the context in encoder input \cite{t5dst},
the decoder generates a sequence consisting of the question and answer: $x = [q; a]$.
We can thus cache the document encoding at inference to answer multiple questions.
At training time we perform next-token prediction, calculating cross-entropy loss by maximizing the log likelihood of the question and answer conditioned on the document: $P(q, a|D) = \prod^T_t P(x_t | x_{<t}, D)$.
We assess the impact of our pre-training on Base ($l$=12, $h$=768) and Large ($l$=24, $h$=1024) models.

\section{Experiments}
\label{sec:exp}

We demonstrate the effectiveness of our pre-training 
on two tasks that require type understanding: zero-shot domain generalization in dialog state tracking (DST), and fine-grained entity typing.


\paragraph{Zero-Shot DST}
\label{dst-exp}
The goal of Dialog State Tracking (DST) is to infer user intent and goals from conversations by filling in belief slots \cite{lemon_dst,wang_dst}.
In many real-world settings, DST models must be able to predict new slot values (i.e.~new entities that are not present in the training corpus) and new slot types (e.g.~requirements for applications in new domains).
This problem setting is known as zero-shot DST (\Cref{tab:dst-example}).
We follow the zero-shot setting in \citet{campagna}: train a model on multi-domain DST data and evaluate on a held-out domain.
We measure domain generalization performance via joint goal accuracy (JGA): the percent of turns in which a model successfully predicts values for all slots in the target domain.
We use the MultiWOZ 2.1 benchmark \cite{multiwoz21},
evaluating zero-shot JGA for the Restaurant, Hotel, Attraction, Train, and Taxi domains.
At each turn, we ask the model a question about the preference for each slot.
We compare against recent systems that can perform zero-shot DST: TRADE \cite{trade_dst}, MA-DST \cite{ma-dst}, SUMBT \cite{sumbt}, and GPT2-DST \cite{gpt2dst}.
Our method is complementary to systems for creating synthetic in-domain dialogs \cite{neuralwoz}.


\begin{table}[t!]
\centering
\small
\begin{tabular}{@{}lrrrrrr@{}}
\toprule
            &\hspace{-12mm} \# Params    & R    & H    & A    & T    & X    \\ \midrule
TRADE       & 90M   & 12.6 & 14.2 & 20.1 & 22.4 & 59.2 \\
MA-DST      & 90M   & 13.6 & 16.3 & 22.5 & 22.8 & 59.3 \\
SUMBT       & 355M  & 16.5 & 19.8 & 22.6 & 22.5 & 59.5 \\
GPT2-DST    & 355M  & 26.2 & 24.4 & 31.3 & 29.1 & 59.6 \\ \midrule
BART        & 139M  & 27.9 & 31.9 & 38.4 & 34.3 & 70.5 \\
Ours (Base) & 139M  & \emph{40.4} & \emph{36.5} & \emph{39.8} & \emph{36.1} & \emph{70.9} \\
Ours (Large)& 406M  & \textbf{46.7} & \textbf{38.8} & \textbf{49.8} & \textbf{37.7}    & \textbf{72.1} \\ \bottomrule
\end{tabular}
\caption{Zero-shot domain adaptation JGA (\%) on MultiWOZ 2.1 test set on the (R)estaurant, (H)otel, (A)ttraction, (T)rain, and Ta(X)i domains. We achieve SOTA results on all domains by significant margins. }
\label{tab:dst}
\end{table}

As seen in \Cref{tab:dst}, \emph{our} type-centric pre-training allows a model to answer questions about unseen slots.
BART-base itself achieves SOTA JGA across all domains, and our pre-training significantly increases the gain to 10.6\% absolute / 34.8\% relative JGA---despite only using one-third of the parameters.
Our Large model achieves 14.9\% absolute and 49.4\% relative gain in JGA compared to previous SOTA.
The most significant gains come in the Hotel and Restaurant domains, which contain the most categorical slots that resemble types (e.g.~cuisine, hotel type).
In \Cref{tab:dst-ablation} we compare our models against same-size BART models at different levels of training data availability to demonstrate the additive utility of our method.
Our method is particularly helpful with less fine-tuning data (low-data regimes), with average gains of 39\% for small models and 4.8\% for large models at 20\% data availability.
Gains are magnified for smaller models, affirming that our method can effectively instill type knowledge in lightweight language models.

\paragraph{Ultra-Fine Entity Typing}
\label{typing-exp}
Our method improves generalization in type-adjacent tasks; we next aim to infer entity types in unseen documents.
In preliminary experiments on the UltraFine dataset with 11K types \cite{ultrafine}, our models under-perform SOTA
(24.0 vs. 49.1 F1).
Manual inspection of gold labels reveals two main causes for error:
1) inaccurate labels---e.g.~``rare plants'' as type ``bird'';
and 2) inconsistent usage of gold labels:
different spellings (\emph{organization} / \emph{organisation}) or synonyms (\emph{car} / \emph{automobile}) are treated as distinct and often do not collocate.
This suggests that label noise in UltraFine may make it unsuitable for assessing granular, hierarchical type knowledge.

We examine these annotation errors via \textbf{human evaluation}, presenting crowd-workers with 200 contexts from UltraFine (10\% of the test set).
Only 68\% of gold type labels were judged accurate, and 21\% inaccurate.
We compare gold labels against zero-shot predictions from our model
in a second trial with 200 pairs.
Judges preferred our predictions 51\% of the time compared to 29\% for gold.
We observed moderate inter-annotator agreement of $\kappa$=0.4044 \cite{fleisskappa}.
This suggests that our models can accurately infer types, but current benchmarks do not suitably measure typing quality.

\begin{table}[t!]
\centering
\small
\begin{tabular}{@{}lrrr@{}}
\toprule
                      & 100\% & 50\%  & 20\% \\ \midrule
Base (139M)           & 13.7  & 14.7  & 39.0 \\
Large (406M)          & 0.9   & 1.6   & 4.8  \\ \bottomrule
\end{tabular}
\caption{Relative gain (\%) in JGA for models trained on WikiWiki vs standard BART pre-training.
Our method helps more in low-data regimes
and for smaller models.}
\label{tab:dst-ablation}
\end{table}



\paragraph{Entity Typing on WikiWiki}
We turn to WikiWiki 
to evaluate fine-grained entity typing, leveraging type labels verified by active users of Wikidata.
To verify the accuracy of ground-truth type labels in the WikiWiki test set, we asked human evaluators to judge the accuracy of 443 type labels from 200 randomly sampled contexts.
We confirm that WikiWiki is a high-quality benchmark for entity typing, with 85\% type precision assessed by human judges (compared to 68\% for UltraFine).

We found that multi-label classifiers built on RoBERTa \cite{roberta} that perform well on UltraFine require significant hyper-parameter tuning to output non-trivial predictions to classify our large and sparse (41K) type ontology.
To perform entity typing with our model, we generate comma-delimited text sequences of types \cite{seq2seqmultilabel}.
This allows our models to infer and generate novel types while classifiers remain restricted to the training ontology.
We confirm that our pre-training helps models better infer types for both seen (+14.3 F1) and unseen entities (+16.7 F1) in new contexts compared to classifiers
(\Cref{tab:typing-automatic}).

\begin{table}[t!]
\centering
\small
\begin{tabular}{@{}llrrr@{}}
\toprule
Entities & Model & \multicolumn{1}{r}{Precision} & \multicolumn{1}{r}{Recall} & \multicolumn{1}{r}{F1} \\ \midrule
Seen   & RoBERTa        & 62.35          & 59.38          & 60.82          \\
       & Ours           & \textbf{78.13} & \textbf{72.39} & \textbf{75.15} \\ \midrule
Unseen & RoBERTa        & 48.88          & 47.96          & 48.41          \\
       & Ours           & \textbf{66.65} & \textbf{63.71} & \textbf{65.14} \\ \bottomrule
\end{tabular}
\caption{P/R/F1 of pred. vs. gold types on WikiWiki Test (seen) and Test New Ent (unseen entities) splits.}
\label{tab:typing-automatic}
\end{table}

To investigate if our model can discover novel types,
we perform another \textbf{human evaluation} over 557 such predictions from 300 contexts, with inter-annotator agreement of $\kappa$=0.4086.
Our model accurately extrapolates its type knowledge beyond the training ontology---we observe 73.3\% precision when inferring new types (compared to 74.5\% precision for seen types),
demonstrating that our pre-training enables models to reason about types beyond simple memorization.
Our model discovers complex and specific scientific types, correctly proposing that anorthosite (an aluminum silicate rock) is a \emph{metallurgical rock}\footnote{rocks containing metallic compounds and properties} and that speckled tortoises are \emph{monotrophs}.\footnote{has diet comprising one type of food \cite{monotroph}}
This reflects the robust taxonomy of types in scientific disciplines.
Our model also proposes granular categories of events, and is judged to correctly type the 2015 Tour of Taiwan as an instance of the \emph{Tour de Taiwan} cycling race.
In the future, we seek methods to automatically assess the factual accuracy of new types.

\section{Conclusion}
In this paper, we 1) propose a text-to-text pre-training scheme to instill type knowledge in language models via QA and 2) release the \textbf{WikiWiki} dataset built from Wikipedia articles and the Wikidata KG.
We show that WikiWiki is larger-scale and more accurate than existing fine-grained type recognition datasets.
We demonstrate that our type-centric pre-training framework allows us to train language models that can better generalize to unseen domains, entities, and types---which in turn lead to improved model performance on downstream tasks like dialog state tracking (achieving SOTA results on zero-shot DST with average gains of 14.9\% joint accuracy).
Our models can extrapolate type knowledge and infer novel types that humans judge to be useful and precise.
As the body of human knowledge grows, we see an opportunity to use life-long learning \cite{lifelong} on news and publications to expand and model the taxonomy of knowledge.

\section*{Acknowledgements}
We would like to thank Stephen Rawls, Ryan Gabbard, and anonymous reviewers for providing valuable feedback on this work.
We also thank Nicolas Gu\'enon des Mesnards and Victor Soto for their help setting up MTurk for human evaluations.
Work was performed during first author’s internship at Amazon Alexa AI.
Findings and observations are of the authors only, and do not necessarily reflect the views of Amazon or UCSD.

\bibliography{anthology,custom}

\begin{thebibliography}{63}
\expandafter\ifx\csname natexlab\endcsname\relax\def\natexlab#1{#1}\fi

\bibitem[{Agarwal et~al.(2021)Agarwal, Ge, Shakeri, and Al{-}Rfou}]{tekgen}
Oshin Agarwal, Heming Ge, Siamak Shakeri, and Rami Al{-}Rfou. 2021.
\newblock \href {https://doi.org/10.18653/v1/2021.naacl-main.278} {Knowledge
  graph based synthetic corpus generation for knowledge-enhanced language model
  pre-training}.
\newblock In \emph{NAACL-HLT}, pages 3554--3565.

\bibitem[{Campagna et~al.(2020)Campagna, Foryciarz, Moradshahi, and
  Lam}]{campagna}
Giovanni Campagna, Agata Foryciarz, Mehrad Moradshahi, and Monica~S. Lam. 2020.
\newblock \href {https://doi.org/10.18653/v1/2020.acl-main.12} {Zero-shot
  transfer learning with synthesized data for multi-domain dialogue state
  tracking}.
\newblock In \emph{ACL}, pages 122--132.

\bibitem[{Cao et~al.(2021)Cao, Izacard, Riedel, and Petroni}]{genre}
Nicola~De Cao, Gautier Izacard, Sebastian Riedel, and Fabio Petroni. 2021.
\newblock \href {https://openreview.net/forum?id=5k8F6UU39V} {Autoregressive
  entity retrieval}.
\newblock In \emph{ICLR}.

\bibitem[{Choi et~al.(2018)Choi, Levy, Choi, and Zettlemoyer}]{ultrafine}
Eunsol Choi, Omer Levy, Yejin Choi, and Luke Zettlemoyer. 2018.
\newblock \href {https://doi.org/10.18653/v1/P18-1009} {Ultra-fine entity
  typing}.
\newblock In \emph{ACL}, pages 87--96.

\bibitem[{Dai et~al.(2021)Dai, Song, and Wang}]{daiET}
Hongliang Dai, Yangqiu Song, and Haixun Wang. 2021.
\newblock \href {https://doi.org/10.18653/v1/2021.acl-long.141} {Ultra-fine
  entity typing with weak supervision from a masked language model}.
\newblock In \emph{ACL/IJCNLP}, pages 1790--1799.

\bibitem[{Dinan et~al.(2019)Dinan, Roller, Shuster, Fan, Auli, and
  Weston}]{wow}
Emily Dinan, Stephen Roller, Kurt Shuster, Angela Fan, Michael Auli, and Jason
  Weston. 2019.
\newblock \href {https://openreview.net/forum?id=r1l73iRqKm} {Wizard of
  wikipedia: Knowledge-powered conversational agents}.
\newblock In \emph{ICLR}. OpenReview.net.

\bibitem[{Eric et~al.(2019)Eric, Goel, Paul, Sethi, Agarwal, Gao, and
  Hakkani{-}T{\"{u}}r}]{multiwoz21}
Mihail Eric, Rahul Goel, Shachi Paul, Abhishek Sethi, Sanchit Agarwal, Shuyang
  Gao, and Dilek Hakkani{-}T{\"{u}}r. 2019.
\newblock \href {http://arxiv.org/abs/1907.01669} {Multiwoz 2.1: Multi-domain
  dialogue state corrections and state tracking baselines}.
\newblock \emph{CoRR}, abs/1907.01669.

\bibitem[{F{\'{e}}vry et~al.(2020{\natexlab{a}})F{\'{e}}vry, Soares,
  FitzGerald, Choi, and Kwiatkowski}]{typesQA}
Thibault F{\'{e}}vry, Livio~Baldini Soares, Nicholas FitzGerald, Eunsol Choi,
  and Tom Kwiatkowski. 2020{\natexlab{a}}.
\newblock \href {https://doi.org/10.18653/v1/2020.emnlp-main.400} {Entities as
  experts: Sparse memory access with entity supervision}.
\newblock In \emph{EMNLP}, pages 4937--4951.

\bibitem[{F{\'{e}}vry et~al.(2020{\natexlab{b}})F{\'{e}}vry, Soares,
  FitzGerald, Choi, and Kwiatkowski}]{EAE}
Thibault F{\'{e}}vry, Livio~Baldini Soares, Nicholas FitzGerald, Eunsol Choi,
  and Tom Kwiatkowski. 2020{\natexlab{b}}.
\newblock \href {https://doi.org/10.18653/v1/2020.emnlp-main.400} {Entities as
  experts: Sparse memory access with entity supervision}.
\newblock In \emph{EMNLP}. Association for Computational Linguistics.

\bibitem[{Fleiss(1971)}]{fleisskappa}
Joseph~L Fleiss. 1971.
\newblock \href
  {http://www.wpic.pitt.edu/research/biometrics/Publications/Biometrics\%20Archives\%20PDF/395-1971\%20Fleiss0001.pdf}
  {Measuring nominal scale agreement among many raters.}
\newblock \emph{Psychological bulletin}, 76(5):378.

\bibitem[{Ganea and Hofmann(2017)}]{ganeaNED}
Octavian{-}Eugen Ganea and Thomas Hofmann. 2017.
\newblock \href {https://doi.org/10.18653/v1/d17-1277} {Deep joint entity
  disambiguation with local neural attention}.
\newblock In \emph{EMNLP}, pages 2619--2629.

\bibitem[{Gehman et~al.(2020)Gehman, Gururangan, Sap, Choi, and
  Smith}]{toxicity}
Samuel Gehman, Suchin Gururangan, Maarten Sap, Yejin Choi, and Noah~A. Smith.
  2020.
\newblock \href {https://doi.org/10.18653/v1/2020.findings-emnlp.301}
  {Realtoxicityprompts: Evaluating neural toxic degeneration in language
  models}.
\newblock In \emph{EMNLP (Findings)}, pages 3356--3369.

\bibitem[{Guu et~al.(2020)Guu, Lee, Tung, Pasupat, and Chang}]{realm}
Kelvin Guu, Kenton Lee, Zora Tung, Panupong Pasupat, and Ming{-}Wei Chang.
  2020.
\newblock \href {http://arxiv.org/abs/2002.08909} {{REALM:} retrieval-augmented
  language model pre-training}.
\newblock \emph{CoRR}, abs/2002.08909.

\bibitem[{Heinzerling and Inui(2021)}]{LMent}
Benjamin Heinzerling and Kentaro Inui. 2021.
\newblock \href {https://aclanthology.org/2021.eacl-main.153/} {Language models
  as knowledge bases: On entity representations, storage capacity, and
  paraphrased queries}.
\newblock In \emph{EACL}, pages 1772--1791.

\bibitem[{Herrera(1976)}]{monotroph}
Carlos~M Herrera. 1976.
\newblock A trophic diversity index for presence-absence food data.
\newblock \emph{Oecologia}, 25(2):187--191.

\bibitem[{Howard and Ruder(2018)}]{ulmfit}
Jeremy Howard and Sebastian Ruder. 2018.
\newblock \href {https://doi.org/10.18653/v1/P18-1031} {Universal language
  model fine-tuning for text classification}.
\newblock In \emph{ACL}, pages 328--339. Association for Computational
  Linguistics.

\bibitem[{Karpukhin et~al.(2020)Karpukhin, Oguz, Min, Lewis, Wu, Edunov, Chen,
  and Yih}]{DPRQA}
Vladimir Karpukhin, Barlas Oguz, Sewon Min, Patrick S.~H. Lewis, Ledell Wu,
  Sergey Edunov, Danqi Chen, and Wen{-}tau Yih. 2020.
\newblock \href {https://doi.org/10.18653/v1/2020.emnlp-main.550} {Dense
  passage retrieval for open-domain question answering}.
\newblock In \emph{EMNLP}, pages 6769--6781.

\bibitem[{Kim et~al.(2021)Kim, Chang, and Lee}]{neuralwoz}
Sungdong Kim, Minsuk Chang, and Sang{-}Woo Lee. 2021.
\newblock \href {https://doi.org/10.18653/v1/2021.acl-long.287} {Neuralwoz:
  Learning to collect task-oriented dialogue via model-based simulation}.
\newblock In \emph{ACL}, pages 3704--3717.

\bibitem[{Kumar et~al.(2020)Kumar, Ku, Goyal, Metallinou, and
  Hakkani{-}T{\"{u}}r}]{ma-dst}
Adarsh Kumar, Peter Ku, Anuj~Kumar Goyal, Angeliki Metallinou, and Dilek
  Hakkani{-}T{\"{u}}r. 2020.
\newblock \href {https://aaai.org/ojs/index.php/AAAI/article/view/6322}
  {{MA-DST:} multi-attention-based scalable dialog state tracking}.
\newblock In \emph{AAAI}, pages 8107--8114.

\bibitem[{Lee et~al.(2019)Lee, Lee, and Kim}]{sumbt}
Hwaran Lee, Jinsik Lee, and Tae{-}Yoon Kim. 2019.
\newblock \href {https://doi.org/10.18653/v1/p19-1546} {{SUMBT:} slot-utterance
  matching for universal and scalable belief tracking}.
\newblock In \emph{ACL}, pages 5478--5483.

\bibitem[{Lemon et~al.(2006)Lemon, Georgila, Henderson, and
  Stuttle}]{lemon_dst}
Oliver Lemon, Kallirroi Georgila, James Henderson, and Matthew~N. Stuttle.
  2006.
\newblock \href {https://www.aclweb.org/anthology/E06-2009/} {An {ISU} dialogue
  system exhibiting reinforcement learning of dialogue policies: Generic
  slot-filling in the {TALK} in-car system}.
\newblock In \emph{EACL}.

\bibitem[{Lewis et~al.(2020{\natexlab{a}})Lewis, Liu, Goyal, Ghazvininejad,
  Mohamed, Levy, Stoyanov, and Zettlemoyer}]{bart}
Mike Lewis, Yinhan Liu, Naman Goyal, Marjan Ghazvininejad, Abdelrahman Mohamed,
  Omer Levy, Veselin Stoyanov, and Luke Zettlemoyer. 2020{\natexlab{a}}.
\newblock \href {https://doi.org/10.18653/v1/2020.acl-main.703} {{BART:}
  denoising sequence-to-sequence pre-training for natural language generation,
  translation, and comprehension}.
\newblock In \emph{ACL}, pages 7871--7880.

\bibitem[{Lewis et~al.(2020{\natexlab{b}})Lewis, Perez, Piktus, Petroni,
  Karpukhin, Goyal, K{\"{u}}ttler, Lewis, Yih, Rockt{\"{a}}schel, Riedel, and
  Kiela}]{rag}
Patrick S.~H. Lewis, Ethan Perez, Aleksandra Piktus, Fabio Petroni, Vladimir
  Karpukhin, Naman Goyal, Heinrich K{\"{u}}ttler, Mike Lewis, Wen{-}tau Yih,
  Tim Rockt{\"{a}}schel, Sebastian Riedel, and Douwe Kiela. 2020{\natexlab{b}}.
\newblock \href
  {https://proceedings.neurips.cc/paper/2020/hash/6b493230205f780e1bc26945df7481e5-Abstract.html}
  {Retrieval-augmented generation for knowledge-intensive {NLP} tasks}.
\newblock In \emph{NeurIPS}.

\bibitem[{Li et~al.(2021)Li, Cao, Sridhar, Zhu, Li, Hamza, and
  McAuley}]{gpt2dst}
Shuyang Li, Jin Cao, Mukund Sridhar, Henghui Zhu, Shang{-}Wen Li, Wael Hamza,
  and Julian~J. McAuley. 2021.
\newblock \href {https://aclanthology.org/2021.eacl-main.91/} {Zero-shot
  generalization in dialog state tracking through generative question
  answering}.
\newblock In \emph{EACL}, pages 1063--1074.

\bibitem[{Lin et~al.(2021)Lin, Liu, Moon, Crook, Zhou, Wang, Yu, Madotto, Cho,
  and Subba}]{t5dst}
Zhaojiang Lin, Bing Liu, Seungwhan Moon, Paul~A. Crook, Zhenpeng Zhou, Zhiguang
  Wang, Zhou Yu, Andrea Madotto, Eunjoon Cho, and Rajen Subba. 2021.
\newblock \href {http://arxiv.org/abs/2105.04222} {Leveraging slot descriptions
  for zero-shot cross-domain dialogue state tracking}.
\newblock \emph{CoRR}, abs/2105.04222.

\bibitem[{Ling and Weld(2012)}]{figer}
Xiao Ling and Daniel~S. Weld. 2012.
\newblock \href {http://www.aaai.org/ocs/index.php/AAAI/AAAI12/paper/view/5152}
  {Fine-grained entity recognition}.
\newblock In \emph{AAAI}. {AAAI} Press.

\bibitem[{Liu et~al.(2020)Liu, Gong, Fu, Yan, Chen, Jiang, Lv, and
  Duan}]{rikinet}
Dayiheng Liu, Yeyun Gong, Jie Fu, Yu~Yan, Jiusheng Chen, Daxin Jiang, Jiancheng
  Lv, and Nan Duan. 2020.
\newblock \href {https://doi.org/10.18653/v1/2020.acl-main.604} {Rikinet:
  Reading wikipedia pages for natural question answering}.
\newblock In \emph{ACL}, pages 6762--6771.

\bibitem[{Liu et~al.(2019)Liu, Ott, Goyal, Du, Joshi, Chen, Levy, Lewis,
  Zettlemoyer, and Stoyanov}]{roberta}
Yinhan Liu, Myle Ott, Naman Goyal, Jingfei Du, Mandar Joshi, Danqi Chen, Omer
  Levy, Mike Lewis, Luke Zettlemoyer, and Veselin Stoyanov. 2019.
\newblock \href {http://arxiv.org/abs/1907.11692} {Roberta: {A} robustly
  optimized {BERT} pretraining approach}.
\newblock \emph{CoRR}, abs/1907.11692.

\bibitem[{{Logan IV} et~al.(2019){Logan IV}, Liu, Peters, Gardner, and
  Singh}]{kglm}
Robert~L. {Logan IV}, Nelson~F. Liu, Matthew~E. Peters, Matt Gardner, and
  Sameer Singh. 2019.
\newblock \href {https://doi.org/10.18653/v1/p19-1598} {Barack's wife hillary:
  Using knowledge graphs for fact-aware language modeling}.
\newblock In \emph{ACL}, pages 5962--5971.

\bibitem[{Lu et~al.(2021)Lu, Lu, Fu, and Liu}]{kelm}
Yinquan Lu, Haonan Lu, Guirong Fu, and Qun Liu. 2021.
\newblock \href {http://arxiv.org/abs/2109.04223} {{KELM:} knowledge enhanced
  pre-trained language representations with message passing on hierarchical
  relational graphs}.
\newblock \emph{CoRR}, abs/2109.04223.

\bibitem[{Lucariello et~al.(1992)Lucariello, Kyratzis, and Nelson}]{cogdev}
Joan Lucariello, Amy Kyratzis, and Katherine Nelson. 1992.
\newblock Taxonomic knowledge: What kind and when?
\newblock \emph{Child development}, 63(4):978--998.

\bibitem[{Majumder et~al.(2020)Majumder, Li, Ni, and McAuley}]{interview}
Bodhisattwa~Prasad Majumder, Shuyang Li, Jianmo Ni, and Julian~J. McAuley.
  2020.
\newblock \href {https://doi.org/10.18653/v1/2020.emnlp-main.653} {Interview:
  Large-scale modeling of media dialog with discourse patterns and knowledge
  grounding}.
\newblock In \emph{EMNLP}, pages 8129--8141.

\bibitem[{Mazar{\'{e}} et~al.(2018)Mazar{\'{e}}, Humeau, Raison, and
  Bordes}]{personachat}
Pierre{-}Emmanuel Mazar{\'{e}}, Samuel Humeau, Martin Raison, and Antoine
  Bordes. 2018.
\newblock \href {https://doi.org/10.18653/v1/d18-1298} {Training millions of
  personalized dialogue agents}.
\newblock In \emph{EMNLP}, pages 2775--2779.

\bibitem[{McCann et~al.(2018)McCann, Keskar, Xiong, and Socher}]{decaNLP}
Bryan McCann, Nitish~Shirish Keskar, Caiming Xiong, and Richard Socher. 2018.
\newblock \href {http://arxiv.org/abs/1806.08730} {The natural language
  decathlon: Multitask learning as question answering}.
\newblock \emph{CoRR}, abs/1806.08730.

\bibitem[{Onoe and Durrett(2020)}]{interpET}
Yasumasa Onoe and Greg Durrett. 2020.
\newblock \href {https://doi.org/10.18653/v1/2020.findings-emnlp.54}
  {Interpretable entity representations through large-scale typing}.
\newblock In \emph{Findings of EMNLP}, pages 612--624.

\bibitem[{Parisi et~al.(2019)Parisi, Kemker, Part, Kanan, and
  Wermter}]{lifelong}
German~Ignacio Parisi, Ronald Kemker, Jose~L. Part, Christopher Kanan, and
  Stefan Wermter. 2019.
\newblock \href {https://doi.org/10.1016/j.neunet.2019.01.012} {Continual
  lifelong learning with neural networks: {A} review}.
\newblock \emph{Neural Networks}, 113:54--71.

\bibitem[{Peters et~al.(2019)Peters, Neumann, IV, Schwartz, Joshi, Singh, and
  Smith}]{knowbert}
Matthew~E. Peters, Mark Neumann, Robert L.~Logan IV, Roy Schwartz, Vidur Joshi,
  Sameer Singh, and Noah~A. Smith. 2019.
\newblock \href {https://doi.org/10.18653/v1/D19-1005} {Knowledge enhanced
  contextual word representations}.
\newblock In \emph{EMNLP}, pages 43--54.

\bibitem[{Petroni et~al.(2019)Petroni, Rockt{\"{a}}schel, Riedel, Lewis,
  Bakhtin, Wu, and Miller}]{LMKG}
Fabio Petroni, Tim Rockt{\"{a}}schel, Sebastian Riedel, Patrick S.~H. Lewis,
  Anton Bakhtin, Yuxiang Wu, and Alexander~H. Miller. 2019.
\newblock \href {https://doi.org/10.18653/v1/D19-1250} {Language models as
  knowledge bases?}
\newblock In \emph{EMNLP}, pages 2463--2473.

\bibitem[{Pudipeddi et~al.(2020)Pudipeddi, Mesmakhosroshahi, Xi, and
  Bharadwaj}]{deepspeed}
Bharadwaj Pudipeddi, Maral Mesmakhosroshahi, Jinwen Xi, and Sujeeth Bharadwaj.
  2020.
\newblock \href {http://arxiv.org/abs/2002.05645} {Training large neural
  networks with constant memory using a new execution algorithm}.
\newblock \emph{CoRR}, abs/2002.05645.

\bibitem[{Qin et~al.(2021)Qin, Lin, Takanobu, Liu, Li, Ji, Huang, Sun, and
  Zhou}]{erica}
Yujia Qin, Yankai Lin, Ryuichi Takanobu, Zhiyuan Liu, Peng Li, Heng Ji, Minlie
  Huang, Maosong Sun, and Jie Zhou. 2021.
\newblock \href {https://doi.org/10.18653/v1/2021.acl-long.260} {{ERICA:}
  improving entity and relation understanding for pre-trained language models
  via contrastive learning}.
\newblock In \emph{ACL}, pages 3350--3363. Association for Computational
  Linguistics.

\bibitem[{Raffel et~al.(2020)Raffel, Shazeer, Roberts, Lee, Narang, Matena,
  Zhou, Li, and Liu}]{t5}
Colin Raffel, Noam Shazeer, Adam Roberts, Katherine Lee, Sharan Narang, Michael
  Matena, Yanqi Zhou, Wei Li, and Peter~J. Liu. 2020.
\newblock \href {http://jmlr.org/papers/v21/20-074.html} {Exploring the limits
  of transfer learning with a unified text-to-text transformer}.
\newblock \emph{JMLR}, 21:140:1--140:67.

\bibitem[{Rastogi et~al.(2020)Rastogi, Zang, Sunkara, Gupta, and
  Khaitan}]{dstc8}
Abhinav Rastogi, Xiaoxue Zang, Srinivas Sunkara, Raghav Gupta, and Pranav
  Khaitan. 2020.
\newblock \href {http://arxiv.org/abs/2002.01359} {Schema-guided dialogue state
  tracking task at {DSTC8}}.
\newblock \emph{CoRR}, abs/2002.01359.

\bibitem[{Sakor et~al.(2020)Sakor, Singh, Patel, and Vidal}]{falcon}
Ahmad Sakor, Kuldeep Singh, Anery Patel, and Maria{-}Esther Vidal. 2020.
\newblock \href {https://doi.org/10.1145/3340531.3412777} {Falcon 2.0: An
  entity and relation linking tool over wikidata}.
\newblock In \emph{CIKM}, pages 3141--3148. {ACM}.

\bibitem[{Shuster et~al.(2021)Shuster, Poff, Chen, Kiela, and
  Weston}]{LMhallucinate}
Kurt Shuster, Spencer Poff, Moya Chen, Douwe Kiela, and Jason Weston. 2021.
\newblock \href {https://aclanthology.org/2021.findings-emnlp.320} {Retrieval
  augmentation reduces hallucination in conversation}.
\newblock In \emph{EMNLP (Findings)}, pages 3784--3803.

\bibitem[{Singh et~al.(2012)Singh, Subramanya, Pereira, and
  McCallum}]{wikilinks}
Sameer Singh, Amarnag Subramanya, Fernando Pereira, and Andrew McCallum. 2012.
\newblock \href
  {https://web.cs.umass.edu/publication/docs/2012/UM-CS-2012-015.pdf}
  {Wikilinks: A large-scale cross-document coreference corpus labeled via links
  to {Wikipedia}}.
\newblock Technical Report UM-CS-2012-015, University of Massachusetts,
  Amherst.

\bibitem[{Sun et~al.(2020{\natexlab{a}})Sun, Shao, Qiu, Guo, Hu, Huang, and
  Zhang}]{colake}
Tianxiang Sun, Yunfan Shao, Xipeng Qiu, Qipeng Guo, Yaru Hu, Xuanjing Huang,
  and Zheng Zhang. 2020{\natexlab{a}}.
\newblock \href {https://doi.org/10.18653/v1/2020.coling-main.327} {Colake:
  Contextualized language and knowledge embedding}.
\newblock In \emph{COLING}. International Committee on Computational
  Linguistics.

\bibitem[{Sun et~al.(2020{\natexlab{b}})Sun, Wang, Li, Feng, Tian, Wu, and
  Wang}]{ernie}
Yu~Sun, Shuohuan Wang, Yu{-}Kun Li, Shikun Feng, Hao Tian, Hua Wu, and Haifeng
  Wang. 2020{\natexlab{b}}.
\newblock \href {https://ojs.aaai.org/index.php/AAAI/article/view/6428}
  {{ERNIE} 2.0: {A} continual pre-training framework for language
  understanding}.
\newblock In \emph{AAAI}. {AAAI} Press.

\bibitem[{Sutskever et~al.(2014)Sutskever, Vinyals, and Le}]{seq2seq}
Ilya Sutskever, Oriol Vinyals, and Quoc~V. Le. 2014.
\newblock \href
  {https://proceedings.neurips.cc/paper/2014/hash/a14ac55a4f27472c5d894ec1c3c743d2-Abstract.html}
  {Sequence to sequence learning with neural networks}.
\newblock In \emph{NIPS}, pages 3104--3112.

\bibitem[{Thirukovalluru et~al.(2021)Thirukovalluru, Sridhar, Thai, Chanumolu,
  Monath, Ananthakrishnan, and McCallum}]{semparse}
Raghuveer Thirukovalluru, Mukund Sridhar, Dung Thai, Shruti Chanumolu, Nicholas
  Monath, Sankaranarayanan Ananthakrishnan, and Andrew McCallum. 2021.
\newblock \href {https://doi.org/10.18653/v1/2021.repl4nlp-1.24} {Knowledge
  informed semantic parsing for conversational question answering}.
\newblock In \emph{RepL4NLP}, pages 231--240, Online.

\bibitem[{Vaswani et~al.(2017)Vaswani, Shazeer, Parmar, Uszkoreit, Jones,
  Gomez, Kaiser, and Polosukhin}]{transformer}
Ashish Vaswani, Noam Shazeer, Niki Parmar, Jakob Uszkoreit, Llion Jones,
  Aidan~N. Gomez, Lukasz Kaiser, and Illia Polosukhin. 2017.
\newblock \href
  {https://proceedings.neurips.cc/paper/2017/hash/3f5ee243547dee91fbd053c1c4a845aa-Abstract.html}
  {Attention is all you need}.
\newblock In \emph{NeurIPS}, pages 5998--6008.

\bibitem[{Vrandecic(2012)}]{Wikidata}
Denny Vrandecic. 2012.
\newblock \href {https://doi.org/10.1145/2187980.2188242} {Wikidata: a new
  platform for collaborative data collection}.
\newblock In \emph{WWW}, pages 1063--1064. {ACM}.

\bibitem[{Wang et~al.(2021)Wang, Gao, Zhu, Zhang, Liu, Li, and Tang}]{kepler}
Xiaozhi Wang, Tianyu Gao, Zhaocheng Zhu, Zhengyan Zhang, Zhiyuan Liu, Juanzi
  Li, and Jian Tang. 2021.
\newblock \href {https://transacl.org/ojs/index.php/tacl/article/view/2447}
  {{KEPLER:} {A} unified model for knowledge embedding and pre-trained language
  representation}.
\newblock \emph{TACL}, 9:176--194.

\bibitem[{Wang and Lemon(2013)}]{wang_dst}
Zhuoran Wang and Oliver Lemon. 2013.
\newblock \href {https://www.aclweb.org/anthology/W13-4067/} {A simple and
  generic belief tracking mechanism for the dialog state tracking challenge: On
  the believability of observed information}.
\newblock In \emph{SIGDIAL}, pages 423--432.

\bibitem[{Weidinger et~al.(2021)Weidinger, Mellor, Rauh, Griffin, Uesato,
  Huang, Cheng, Glaese, Balle, Kasirzadeh, Kenton, Brown, Hawkins, Stepleton,
  Biles, Birhane, Haas, Rimell, Hendricks, Isaac, Legassick, Irving, and
  Gabriel}]{ethics_gen}
Laura Weidinger, John Mellor, Maribeth Rauh, Conor Griffin, Jonathan Uesato,
  Po{-}Sen Huang, Myra Cheng, Mia Glaese, Borja Balle, Atoosa Kasirzadeh, Zac
  Kenton, Sasha Brown, Will Hawkins, Tom Stepleton, Courtney Biles, Abeba
  Birhane, Julia Haas, Laura Rimell, Lisa~Anne Hendricks, William~S. Isaac,
  Sean Legassick, Geoffrey Irving, and Iason Gabriel. 2021.
\newblock \href {http://arxiv.org/abs/2112.04359} {Ethical and social risks of
  harm from language models}.
\newblock \emph{CoRR}, abs/2112.04359.

\bibitem[{Wu et~al.(2019)Wu, Madotto, Hosseini{-}Asl, Xiong, Socher, and
  Fung}]{trade_dst}
Chien{-}Sheng Wu, Andrea Madotto, Ehsan Hosseini{-}Asl, Caiming Xiong, Richard
  Socher, and Pascale Fung. 2019.
\newblock \href {https://doi.org/10.18653/v1/p19-1078} {Transferable
  multi-domain state generator for task-oriented dialogue systems}.
\newblock In \emph{ACL}, pages 808--819.

\bibitem[{Wu et~al.(2020{\natexlab{a}})Wu, Petroni, Josifoski, Riedel, and
  Zettlemoyer}]{ZSEL}
Ledell Wu, Fabio Petroni, Martin Josifoski, Sebastian Riedel, and Luke
  Zettlemoyer. 2020{\natexlab{a}}.
\newblock \href {https://doi.org/10.18653/v1/2020.emnlp-main.519} {Scalable
  zero-shot entity linking with dense entity retrieval}.
\newblock In \emph{EMNLP}, pages 6397--6407.

\bibitem[{Wu et~al.(2020{\natexlab{b}})Wu, Petroni, Josifoski, Riedel, and
  Zettlemoyer}]{blink}
Ledell Wu, Fabio Petroni, Martin Josifoski, Sebastian Riedel, and Luke
  Zettlemoyer. 2020{\natexlab{b}}.
\newblock \href {https://doi.org/10.18653/v1/2020.emnlp-main.519} {Scalable
  zero-shot entity linking with dense entity retrieval}.
\newblock In \emph{EMNLP}, pages 6397--6407.

\bibitem[{Yamada et~al.(2020)Yamada, Asai, Shindo, Takeda, and
  Matsumoto}]{luke}
Ikuya Yamada, Akari Asai, Hiroyuki Shindo, Hideaki Takeda, and Yuji Matsumoto.
  2020.
\newblock \href {https://doi.org/10.18653/v1/2020.emnlp-main.523} {{LUKE:} deep
  contextualized entity representations with entity-aware self-attention}.
\newblock In \emph{EMNLP}. Association for Computational Linguistics.

\bibitem[{Yamada et~al.(2019)Yamada, Washio, Shindo, and Matsumoto}]{yamadaED}
Ikuya Yamada, Koki Washio, Hiroyuki Shindo, and Yuji Matsumoto. 2019.
\newblock \href {http://arxiv.org/abs/1909.00426} {Global entity disambiguation
  with pretrained contextualized embeddings of words and entities}.
\newblock \emph{CoRR}, abs/1909.00426.

\bibitem[{Yang et~al.(2018)Yang, Sun, Li, Ma, Wu, and Wang}]{seq2seqmultilabel}
Pengcheng Yang, Xu~Sun, Wei Li, Shuming Ma, Wei Wu, and Houfeng Wang. 2018.
\newblock \href {https://aclanthology.org/C18-1330/} {{SGM:} sequence
  generation model for multi-label classification}.
\newblock In \emph{COLING}, pages 3915--3926.

\bibitem[{Yao et~al.(2019)Yao, Ye, Li, Han, Lin, Liu, Liu, Huang, Zhou, and
  Sun}]{docred}
Yuan Yao, Deming Ye, Peng Li, Xu~Han, Yankai Lin, Zhenghao Liu, Zhiyuan Liu,
  Lixin Huang, Jie Zhou, and Maosong Sun. 2019.
\newblock \href {https://doi.org/10.18653/v1/p19-1074} {Docred: {A} large-scale
  document-level relation extraction dataset}.
\newblock In \emph{ACL}, pages 764--777.

\bibitem[{You et~al.(2020)You, Li, Reddi, Hseu, Kumar, Bhojanapalli, Song,
  Demmel, Keutzer, and Hsieh}]{lamb}
Yang You, Jing Li, Sashank~J. Reddi, Jonathan Hseu, Sanjiv Kumar, Srinadh
  Bhojanapalli, Xiaodan Song, James Demmel, Kurt Keutzer, and Cho{-}Jui Hsieh.
  2020.
\newblock \href {https://openreview.net/forum?id=Syx4wnEtvH} {Large batch
  optimization for deep learning: Training {BERT} in 76 minutes}.
\newblock In \emph{ICLR}. OpenReview.net.

\bibitem[{Zhang et~al.(2021)Zhang, Wang, Hu, Qiu, Tang, He, and Huang}]{dkplm}
Taolin Zhang, Chengyu Wang, Nan Hu, Minghui Qiu, Chengguang Tang, Xiaofeng He,
  and Jun Huang. 2021.
\newblock \href {http://arxiv.org/abs/2112.01047} {{DKPLM:} decomposable
  knowledge-enhanced pre-trained language model for natural language
  understanding}.
\newblock \emph{CoRR}, abs/2112.01047.

\end{thebibliography}
\bibliographystyle{acl_natbib}

\newpage
\appendix

\section{Data}
\label{sec:appendix-data}

We use the June 2021 Wikidata database file from \url{https://www.wikidata.org/wiki/Wikidata:Database_download} for raw KG data.
We use English Wikipedia article HTML crawled from the same time period.
While Wikidata contains multilingual definitions and labels for each node, in this paper we use only English entity and type names.

Wikipedia data was collected under the original terms of release which allow free usage of such materials for non-commercial purposes.\footnote{\url{https://en.wikipedia.org/wiki/Wikipedia:Copyrights}}
We will release WikiWiki under the same license.

When creating questions for pre-training tasks, if a question has multiple answers (e.g.~multiple chemists in \Cref{tab:pretrain-ex}), the answers are a comma- and \texttt{and}-delimited sequence, in order of appearance in the context.
For the entity typing question, we use the order that types appear in the Wikidata page.

\section{Experimental Settings}
\label{sec:appendix}

We train all of our models on a node with eight Nvidia V100 GPUs (comprising 256 GB total VRAM) and 768 GB of RAM.
We optimize using Deepspeed Stage 1 \cite{deepspeed} using FP16 and the Lamb optimizer \cite{lamb}.
Experimental results, where applicable, are reported as median of 3 experiments.

\paragraph{Hyperparameters}
For pre-training, we use a learning rate of 1e-4 with a linear warm-up for the first 10\% of training iterations, using an effective batch size of 960.
Our models were trained on a single pass of our pre-training dataset of 50M questions, totaling 52K steps.
We fine-tune models using the same learning rate schedule, using an effective batch size of 2560 and early stopping for a maximum of 10 epochs based on validation loss.
We aim to establish the general ability of our pre-training scheme to instill type awareness, and thus fix hyperparameters for generative language models trained with our method without hyperparameter tuning.

As mentioned in \Cref{sec:exp}, the RoBERTa-based classifier for entity typing on WikiWiki required significantly more hyperparameter tuning; we performed a hyperparameter sweep on batch size (512 to 2048), learning rate (1e-3 to 1e-5), optimizer (Adam vs. Lamb), and whether to freeze the encoder.
We achieved best performance (as in \Cref{tab:typing-automatic}) with a learning rate of 1e-4, the Adam optimizer, an effective batch size of 960, and with gradual unfreezing \cite{ulmfit} over 5K steps.
We found gradual unfreezing to be critical for model performance, with fully frozen and fully unfrozen RoBERTa models achieving entity typing F1 scores of $\leq$ 10.0.

\begin{table}[t!]
\centering
\small
\resizebox{\linewidth}{!}{
\begin{tabular}{@{}lrrrrrr@{}}
\toprule
            &\hspace{-14mm} \# Params    & R    & H    & A    & T    & X    \\ \midrule
GPT2-DST    & 355M  & 26.2 & 24.4 & 31.3 & 29.1 & 59.6 \\
\ \ + SGD   & 355M  & 27.7    & 24.9    & \emph{42.4}    & \textbf{41.1} & 60.3    \\ \midrule
Ours (Base) & 139M  & \emph{40.4} & \emph{36.5} & 39.8 & 36.1 & \emph{ 70.9} \\
Ours (Large)& 406M  & \textbf{46.7} & \textbf{38.8} & \textbf{49.8} & \emph{37.7}    & \textbf{72.1} \\ \bottomrule
\end{tabular}
}
\caption{Zero-shot domain adaptation JGA (\%) on MultiWOZ 2.1 test set on the (R)estaurant, (H)otel, (A)ttraction, (T)rain, and Ta(X)i domains. Compared to GPT2-DST \cite{gpt2dst} augmented with out-of-domain DST data (+SGD), our Base model out-performs the augmented model in 3/5 domains and our Large model out-performs it in 4/5 domains.}
\label{tab:dst-appendix-SGD}
\end{table}

\section{Dialog State Tracking Notes}
As discussed in \Cref{sec:exp}, our method is orthogonal to and thus can be used simultaneously with techniques for creating synthetic in-domain training data for DST \cite{campagna,neuralwoz}.
For slot queries, we use templated questions of the form:
\texttt{What [domain] [slot] is the user interested in?}.

We compare our models against SOTA models for zero-shot DST on MultiWOZ 2.1.
We affirm the observations of \citet{t5dst} that while T5-DST achieves strong DST performance on the 2.0 version of the dataset, performance degrades on the 2.1 benchmark.

\citet{gpt2dst} also present results for GPT2-DST when training is augmented with additional DST data from a wider pool of domains---the Schema-Guided Dialog dataset \cite{dstc8}.
In the interest of fairness, we do not compare this setting in \Cref{tab:dst} as our models do not have access to \emph{any} conversational data in pre-training and---like the other baseline models---cannot access additional DST data in fine-tuning.
Despite the lack of exposure to conversational data, in \Cref{tab:dst-appendix-SGD} we show that our Small and Large models out-perform GPT2-DST + SGD in 3/5 domains (with absolute per-domain gain of 5.5\% and relative gain of 18.3\%) and 4/5 domains (with absolute gains of 9.7\% and relative gains of 30.6\%), respectively.
We additionally present zero-shot DST performance (JGA) on the MultiWOZ 2.1 validation set in \Cref{tab:dst-appendix-validation}.

\begin{table}[t!]
\centering
\small
\begin{tabular}{@{}lrrrrrr@{}}
\toprule
            &\hspace{-12mm} \# Params    & R    & H    & A    & T    & X    \\ \midrule
BART-base   & 139M  & 29.6 & 31.5 & 38.7 & 35.0 & 70.5 \\
Ours (Base) & 139M  & 41.3 & 33.6 & 42.5 & 36.6 & 71.9 \\
Ours (Large)& 406M  & 46.4 & 37.6 & 52.3 & 38.0 & 72.1 \\ \bottomrule
\end{tabular}
\caption{
Zero-shot domain adaptation JGA (\%) on MultiWOZ 2.1 \emph{\textbf{validation}} set on the (R)estaurant, (H)otel, (A)ttraction, (T)rain, and Ta(X)i domains.
}
\label{tab:dst-appendix-validation}
\end{table}

\section{Human Evaluation Details}

We perform our evaluation using the Amazon Mechanical Turk platform.\footnote{\url{https://www.mturk.com/}}
To ensure high quality annotations, we recruit only crowd workers with Master qualification---indicating a history of high quality accepted work---and who are native English speakers.\footnote{\url{https://www.mturk.com/worker/help}}
Crowd-workers remained anonymous outside of their qualifications and we did not collect any additional demographic information.
Workers were informed that their type accuracy judgements were to be used in an academic research setting, with an option to opt-out and reject the task.

As both gold types and predicted types could be complex and require domain knowledge, evaluators were instructed to search any relevant additional material (textbooks, sites, papers) to ensure they made a high confidence judgment of type accuracy.
Based on the average time spent evaluating each article, our pay rate worked out to above Federal minimum wage in the United States.

\begin{figure*}[t!]
    \centering
    \includegraphics[width=1.0\linewidth]{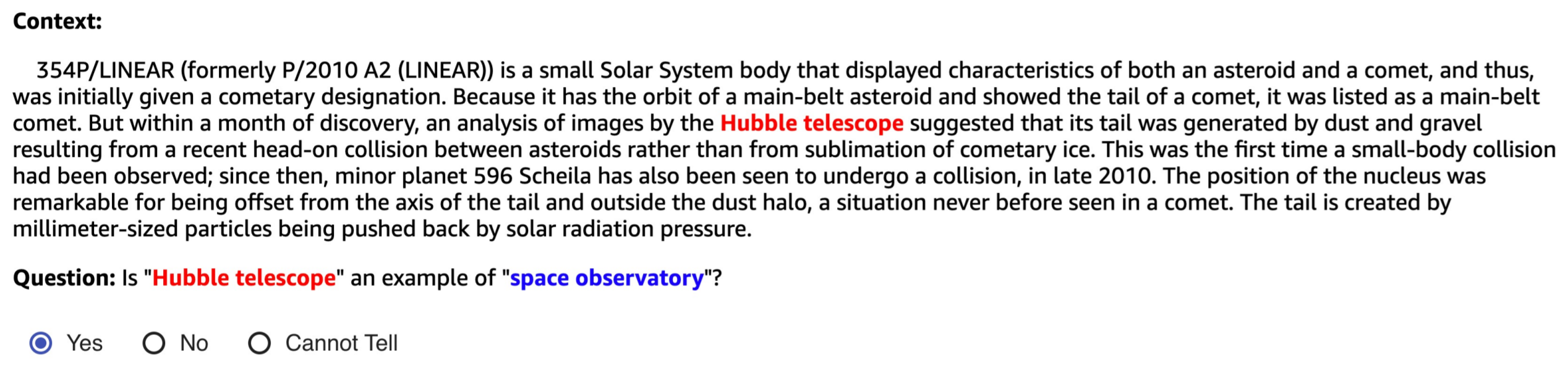}
    \caption{Example of human evaluation question where the judge is asked to assess whether a predicted / ground truth type accurately applies to the entity referenced.}
    \label{fig:he-acc}
\end{figure*}

\begin{figure*}[t!]
    \centering
    \includegraphics[width=1.0\linewidth]{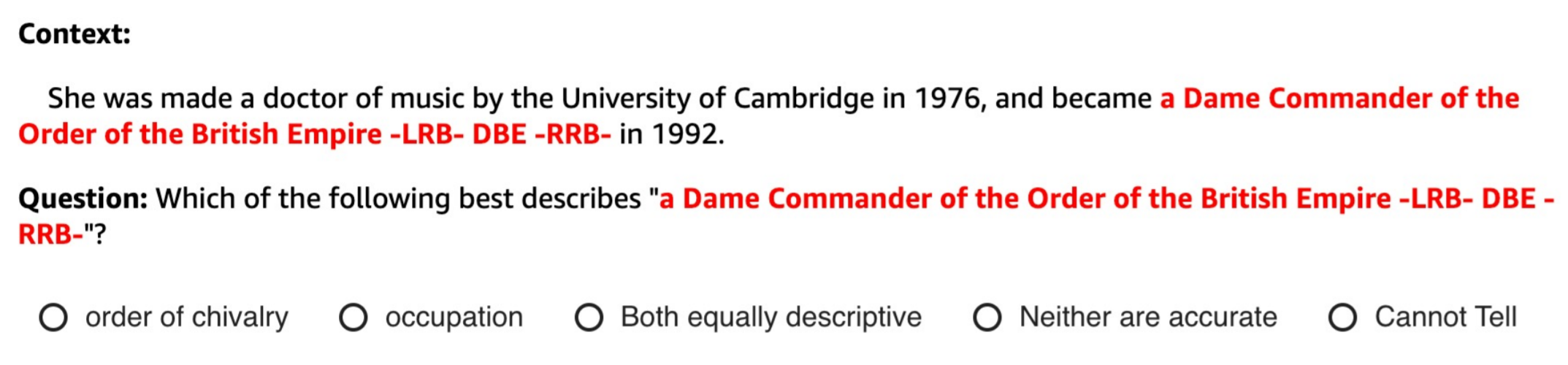}
    \caption{Example of human evaluation question where the judge is asked to assess to determine the relative suitability and quality of two different types for the entity referenced.}
    \label{fig:he-comp}
\end{figure*}

In \Cref{fig:he-acc} we display the example instructions given to a human evaluator for assessing the accuracy of a type for an entity referenced in a context.
In \Cref{fig:he-comp} we show sample instructions given to a human evaluator to choose which of two types (predicted or gold label in random order) is more suitable / applies more accurately to the referenced entity.

\section{Ethics}
As with all models capable of generating arbitrary text sequences, models trained with our framework and tasks run the risk of outputting toxic or offensive text \cite{toxicity}.
However, our training aims to instill type knowledge for type- and concept-reliant downstream tasks.
As such, we expect that our pre-training does not heighten the risk of offensive outputs compared to other general-purpose pre-training schemes on wide internet corpora.

The primary risk of instilling models with type knowledge lies in the potential for misinformation \cite{ethics_gen}.
For example, if our model is used to extend existing taxonomies, it runs the risk of hallucinating false types.
We observe in \Cref{tab:typing-automatic} that while our model achieves high typing precision and recall for seen and unseen types in new documents, we are not at the point where it can be used in isolation to discover and add knowledge to existing knowledge graphs.
In parallel with developing better methods for verifying type ontologies and assignments, it is important to incorporate domain experts or crowd-source verification when language models are used to discover facts or type relationships in new documents.

We also advocate for more careful inspection of racial, gender, and socioeconomic biases in existing type ontologies, as it is possible for type-aware models to propagate such biases (e.g.~associating people with certain patterns of names with specific occupations).

\end{document}